\documentclass{article}

\PassOptionsToPackage{numbers, compress}{natbib}
\usepackage[preprint]{neurips_2022}

\usepackage[utf8]{inputenc} % allow utf-8 input
\usepackage[T1]{fontenc}    % use 8-bit T1 fonts
\usepackage{hyperref}       % hyperlinks
\usepackage{url}            % simple URL typesetting
\usepackage{booktabs}       % professional-quality tables
\usepackage{amsfonts}       % blackboard math symbols
\usepackage{nicefrac}       % compact symbols for 1/2, etc.
\usepackage{microtype}      % microtypography
\usepackage{xcolor}         % colors
\usepackage{graphicx}
\usepackage{amsmath,amssymb}
\usepackage{algorithm}
\usepackage{algorithmic}
\usepackage{multirow}
\usepackage{subfigure}
\usepackage{arydshln}
\usepackage{multirow}
\usepackage{setspace}

\title{Asymptotic Soft Cluster Pruning for Deep Neural Networks}

\author{Tao Niu$^{1}$, Yinglei Teng$^{1}$, Panpan Zou$^{1}$\\ 
	$^1$  Beijing University of Posts and Telecommunications, Beijing, China\\					
	{\tt \scriptsize \{tasakim, lilytengtt, zoupanpan\}@bupt.edu.cn}
}
\begin{document}

\maketitle

\begin{abstract}
Filter pruning method introduces structural sparsity by removing selected filters and is thus particularly effective for reducing complexity. Previous works empirically prune networks from the point of view that filter with smaller norm contributes less to the final results. However, such criteria has been proven sensitive to the distribution of filters, and the accuracy may hard to recover since the capacity gap is fixed once pruned. In this paper, we propose a novel filter pruning method called Asymptotic Soft Cluster Pruning (ASCP), to identify the redundancy of network based on the similarity of filters. Each filter from over-parameterized network is first distinguished by clustering, and then reconstructed to manually introduce redundancy into it. Several guidelines of clustering are proposed to better preserve feature extraction ability. After reconstruction, filters are allowed to be updated to eliminate the effect caused by mistakenly selected. Besides, various decaying strategies of the pruning rate are adopted to stabilize the pruning process and improve the final performance as well. By gradually generating more identical filters within each cluster, ASCP can remove them through channel addition operation with almost no accuracy drop. Extensive experiments on CIFAR-10 and ImageNet datasets show that our method can achieve competitive results compared with many state-of-the-art algorithms. 
%   Our code is available at
\end{abstract}

\section{Introduction}
Deep convolutional neural networks (CNNs) have shown superior performance in a variety of tasks such as computer vision \cite{ref1}, natural language processing \cite{ref2}, and object detection \cite{ref3}. However, its state-of-the-art performance is based on deeper and wider architecture, hindering the deployment in resources-limited devices due to its expensive computation cost and memory footprint. To solve this realistic problem, network pruning has been well studied by a large amount of researchers. Existing solutions on pruning can be roughly divided into two categories: weight pruning and filter pruning. Weight pruning \cite{ref4} \cite{ref5} \cite{ref6} refers to pruning the weights of a filter. However, such fine-grained method will cause unstructured sparsity, which may still be less efficient in saving the memory usage and computational cost, since the unstructured model cannot leverage the existing high-efficiency BLAS libraries. In contrast, filter pruning directly deletes the whole filter and is believed easy to achieve acceleration on general platforms. 

Most of the previous filer pruning methods \cite{ref7} \cite{ref8} \cite{ref9} follow the three-stage pipeline: training a large model, pruning filters via a pre-defined criterion, e.g., $\ell _p$-norm, fine-tuning to compensate for the accuracy loss. The norm-based criterion is based on the assumption that the smaller norm of filter is, the less contribution to final result it brings. One drawback of such pipeline is that it is very sensitive to the distribution of filters. Although the norm of filters is small, it may still contain semantic information which is helpful for further processing. Another problem is that once pruned, there will be a capacity gap compared to the original network. This may make the network performance hard to recover even after fine-tuning. Although methods like Soft Filter Pruning (SFP) \cite{ref10} and SofteR Filter Pruning (SRFP) \cite{ref11} can eliminate this disadvantage by zeroing filters instead of directly removing them, there will be a severe accuracy drop after pruning when the pruning rate is large since they manually force the filters not to contribute at all, which can be seen as strong regularization.

\begin{figure}[t]
\centerline{
\subfigure[Similar feature map pairs]{\includegraphics[width=0.495\textwidth]{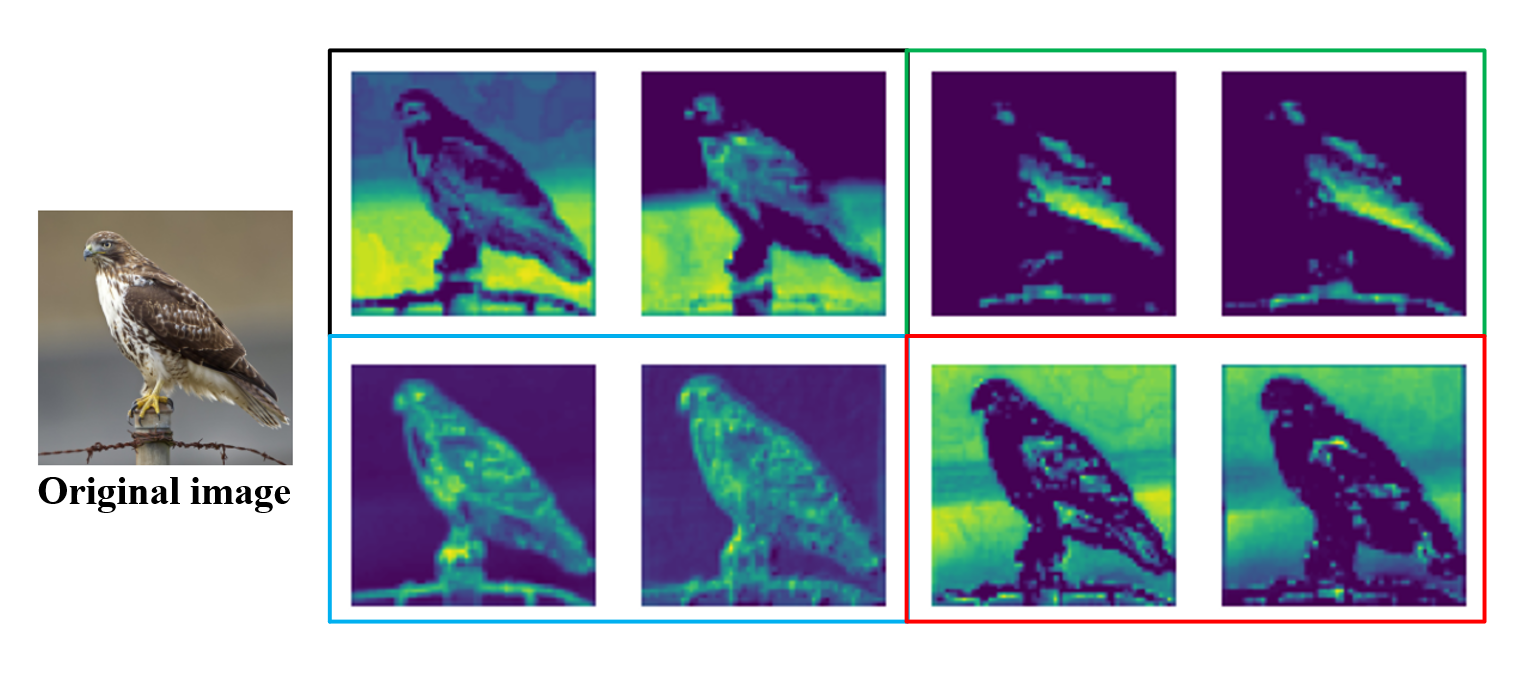}}
\subfigure[Filter distribution of ResNet-50]{\includegraphics[width=0.495\textwidth]{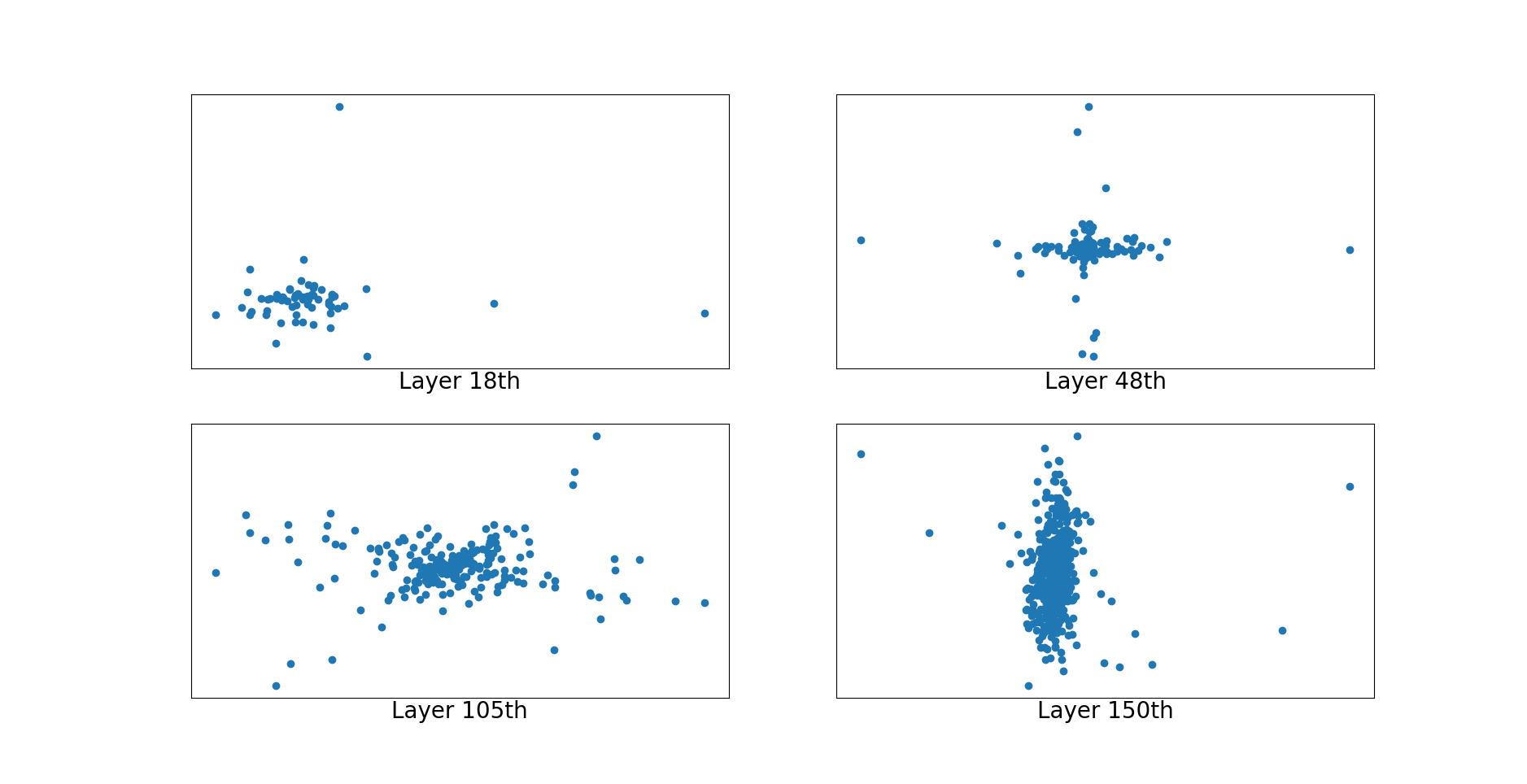}} }
\caption{(a) Similar feature map pairs of the second residual block in ResNet-50. Different pairs are framed in different colors. (b) Visualization of filter distribution of ResNet-50 after decomposition.}
\label{fig1}
\end{figure}

“More-similar-less-important” is another viewpoint to identify pruned filters regardless of their norm. As shown in Fig. \ref{fig1}(a), there are many feature maps pairs of convolutional layers that contain similar semantic information. Such observation has also been pointed out by GhostNet \cite{ref12}, indicating the redundancy is very prevalent on the feature map level, which reveals the similarity of corresponding filters. For further investigation, filters from different depth of the network after decomposition by PCA are visualized in Fig. \ref{fig1}(b), which shows the phenomenon of closing to each other and forming clusters. Although methods like FPGM \cite{ref13} and FPKM \cite{ref14} prune the network from this perspective, setting redundant filter to zero will undoubtedly bring information loss, since similar filters does not mean zero contribution.

To address the problems of distribution sensitive and semantic information loss that mentioned above, in this paper, we propose a novel model pruning method based on clustering. Instead of zeroing out unimportant filters, we intend to set them as value of cluster centroids to preserve their ability of extracting features while training. After convergence, we obtain a network whose filters are divided into several clusters and exactly identical within each cluster. Due to the linear and combinational properties of convolutional layer, we can then discard all but to keep one filter of each clusters of current layer and add up the parameters along the corresponding channels of the next layer. Pruning such way will cause almost zero performance loss, thus allowing to retrain for a very limited epochs to overcome the effect caused by batch normalization, which avoids the time-consuming fine-tuning process in previous works.

To summarize, our main contributions are three-fold as follows:
\vspace{-1ex}
\begin{itemize}
\item We empirically demonstrate the redundancy of filters through similarity of feature maps and propose the Asymptotic Soft Cluster Pruning (ASCP) to generate identical filters in a soft manner. Multiple guidelines of clustering are given through analysis and can theoretically well protect feature extraction ability of filters. 
\end{itemize}

\vspace{-2ex}
\begin{itemize}
\item To prevent from strong regularization, we adopt different strategies of pruning rate to prune in an asymptotic way. After convergence, the channel-wise addition operation is introduced to remove identical filters without information loss, since the compact network is equivalent to the unpruned one regardless of batch normalization, which suggest our method can omit the time-consuming fine-tuning step.
\end{itemize}

\vspace{-2ex}
\begin{itemize}
\item Extensive experiments on CIFAR-10 and ImageNet datasets demonstrate the effectiveness and efficiency of the proposed ASCP compared with other state-of-the-art methods.
\end{itemize}

\section{Related Work}
\label{sec:II}
In this section, we will give a comprehensive overview of different pruning methods, which can be roughly divided into three aspects, pruning structure, pruning criteria and pruning manner.

\begin{figure*}[t]
\centerline{\includegraphics[width=5.5in]{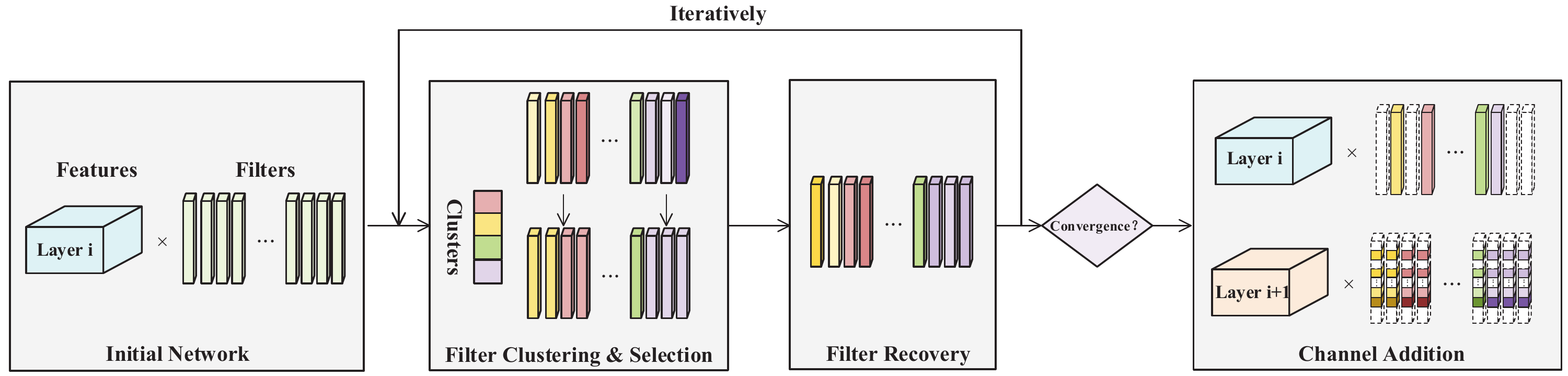}}
\caption{Overview of our ASCP. Filters are clustered and manually reconstructed in the filter clustering and selection step, and are retrained in the filter recovery step. This pipeline is executed iteratively till convergence before channel operation operation is conducted to obtain the compact model. Filters in similar colors are considered close to each other in Euclidean space. (Best viewed in color.)}
\vspace{-1ex}
\label{fig2}
\end{figure*}

\textbf{Pruning Structure.} Early works of network pruning concentrate on fine-grained granularities: weight-level \cite{ref6}, vector-level \cite{ref18} and kernel level \cite{ref15}. For the architecture consistency, this type of non-structured methods can only zeroize the unnecessary parameters rather than pruned. The need of saving special coordinate for every weight makes it difficult to satisfy in nowadays trillion-level model. During inference, although the model sizes and number of multiply-accumulate operations are dramatically decreased, the irregular structure of dense matrices requires additional computations and special hardware designs for acceleration. In line with our work, a few methods have been recently proposed for filter-level pruning (i.e., structured pruning), which can reduce both network size and inference speed and are well supported by various off-the-shelf deep learning libraries. By directly deleting redundant filters in current layer, the corresponding kernels in next layer should also be pruned to keep the consistency of architecture.

\textbf{Pruning Criteria.} Selecting a proper criterion to identify filters which need to be pruned is a major point of network pruning. %Prior works pay most attention to the “smaller-norm-less-important” assumption, where the filters with smaller norms are expected to make absolutely small contributions to the final results.
\cite{ref15} prunes filters with small $\ell _1$-norm in each layer, while \cite{ref19} \cite{ref20} are based on $\ell _2$-norm. \cite{ref21} uses the Taylor series to estimate the loss change after each filter's removal and prune the filters that cause minimal training loss change. Network slimming \cite{ref22} applies LASSO on the scaling factors of BN, by setting the BN scaling factor to zero, channel-wise pruning is enabled. FPGM \cite{ref13} prunes filters nearest to the geometric median of each layer using Euclidian distance, refraining from the limits of norm-based criterion. \cite{ref23} indicates activation may also be an indicator, and it introduces Average Percentage Of Zeros (APoZ) to judge if one output activation map is contributing to the result.

\textbf{Pruning Manner.} The traditional hard filter pruning pipeline will reduce the model capacity of the original models, thus facing the problem of unrecoverable performance loss after incorrect pruning. Besides, the overreliance on pre-trained models and huge time cost of fine-tuning make it unsuitable to deploy in real world scenarios. To overcome that, SFP \cite{ref10} zeroizes pruned filters with a binary mask and updates filters in a soft manner to maintain the capacity of the network. Based on SFP, ASFP \cite{ref24} gradually increases the pruning rate to alleviate the accuracy drop caused by pruning and stabilize the whole pruning process. STP \cite{ref25} uses first-order Taylor series to measure the importance of filters while SRFP \cite{ref11} removes filters smoothly by gradually decaying to zero, so that it can better preserve the trained information. \cite{ref26} analysis the characteristic of hard manner and soft manner, and propose GHFP to smoothly switch from soft to hard to achieve a balance between performance and convergence speed. 
%Inspired by such soft-manner pruning methods, we propose to prune filters in our ASCP in another point of view instead of zeroizing them. 

\section{Methodology}

\subsection{Formulation}

In this section, we formally introduce the symbol and notation. The deep CNN network can be parameterized by $W_i\in \mathbb{R} ^{N_{out}^{i}*N_{in}^{i}*h_i*w_i}$, where $1\leqslant i\leqslant L$, and $L$ is the total number of convolutional layers in a network, $h_i$ and $w_i$ represent the height and width of a kernel. $N_{out}^{i}$ and $N_{in}^{i}$ denote the number of output channels and input channels for the $i$-th convolution layer, respectively. The output feature map $O_i\in \mathbb{R} ^{N_{out}^{i}*h_{i+1}*w_{i+1}}$ is calculated by the convolution operation of input feature map $I_i\in \mathbb{R}^{N_{in}^{i}*h_i*w_i}$ and $W_i$, shown as below:

\begin{equation}
{O_i=W_i*I_i \label{eq1}}.
\end{equation}

$O_{i,j}$ and $W_{i,j}$ denote the $j$-th output channel of output feature maps and the $j$-th filter of the $i$-th layer, respectively. Assume that the pruning rate of $i$-th layer is $P_i$, then the number of filters after pruning will reduce from $N_{out}^{i}$ to $N_{out}^{i}*\left( 1-P_i \right)$. Accordingly,  the size of the pruned output tensor would be $N_{out}^{i}*\left( 1-P_i \right) *h_{i+1}*w_{i+1}$. Given a dataset $D=\left\{ \left( x_i,y_i \right) \right\} _{i=1}^{n}$ and a desired sparsity level $k$ (i.e., the number of remaining filters), hard filter pruning can be formulated as:

\begin{align}\label{eq2}
\mathop {\min } \limits_{W}\ell \left( W^{'};D \right) &=\min_W \frac{1}{n}\sum_{i=1}^n{\ell \left( W^{'};\left( x_i,y_i \right) \right)},
\\
\text{s.t.} \quad \left\| \left. W^{'} \right\| \right. _0 &\leqslant k \notag,
\end{align}

where $\ell(.)$ is the loss function (e.g., cross-entropy loss), $W^{'}$ is the filter set of the pruned network. Typically, hard filter pruning prunes the model layer by layer and fine-tune iteratively to complement the degradation of the performance. However, once filters are pruned, they will not be updated again which may cause capacity reduction. To alleviate that, soft filter pruning (SFP) and its variants prune filters by simply zeroizing them, which can be represented by 

\begin{equation}
{W_{i}^{'}=W_i\odot M_i \label{eq3}},
\end{equation}
where $M_i$ is a Boolean matrix with the same shape as $W_i$, indicates whether the filter is pruned or not. $\odot$ denotes the Hadamard product. Specifically, $M_{i,j}=0$ if $W_{i,j}$
is pruned, otherwise we set $M_{i,j}=1$.

\subsection{Asymptotic Soft Cluster Pruning}
Base on the observation that filters are naturally united in each layer, our ASCP intend to manually introduce similar filters to generate redundancy. As depicted in Fig. \ref{fig2}, our pipeline is carried out iteratively till convergence, and then prunes the network with channel addition operation.

\textbf{Filter Clustering:} We first reshape 4D tensor $W_i$ into a 2D matrix of size ${(N_{in}^{i}*h_{i}*w_{i})*N_{out}^{i}}$. Therefore, each column of this 2D matrix stands for the filter of the weight tensor. Then we can regard them as coordinates in high dimensional space and conduct clustering as:
 
\begin{equation}
{\min \frac{1}{N_{out}^{i}}\sum_{j=1}^{N_{out}^{i}}{\left\| \left. W_{ij}-\mu _{c_{i}^{\left( j \right)}} \right\| \right. _2} \label{eq4}},
\end{equation} 
where $c_i$ is the amount of clusters of $i$-th layer, $\mu _{c_{i}^{\left( j \right)}}$ is the cluster centroid which $W_{ij}$ is assigned to. The optimization can be solved by $kmeans$ \cite{ref27} or other clustering algorithms. After clustering, filters in the $i$-th layer can be divided into $c_i$ clusters, where their norm are relatively close and may have a certain linear relationship \cite{ref13}. Filters within each cluster can generate similar feature maps of the next layer, therefore, the clusters with more filters can be identified as more redundant.
Setting the hyper-parameter $c_i$ properly is quite critical, since it determines the performance of network after pruning. Next, we propose three guidelines for effectively setting. \\
\emph{\textbf{Guideline 1: Silhouette Coefficient.}} With the common used Silhouette Coefficient, we can set $c_i$ by solving the following optimization:

\begin{align}\label{eq5}
c_{i}^{*}&=\mathop {arg\max} \limits_{c_i}\,\,s_i,
\\
\text{where} \quad s_i&=\frac{1}{N_{out}^{i}}\sum_{j=1}^{N_{out}^{i}}{\frac{b_j-a_j}{\max \left( b_j,a_j \right)}} \notag,
\end{align} 
  
where ${a_j}$ denotes the average intra-cluster distance (i.e., the average distance between $W_{ij}$ within a cluster), and ${b_j}$ is the average inter-cluster distance, which means the average distance between $W_{ij}$ and filters within all other clusters. 

\begin{figure}[t]
\centering
\subfigure[Adaptive number of clusters]{\includegraphics[width=0.495\textwidth]{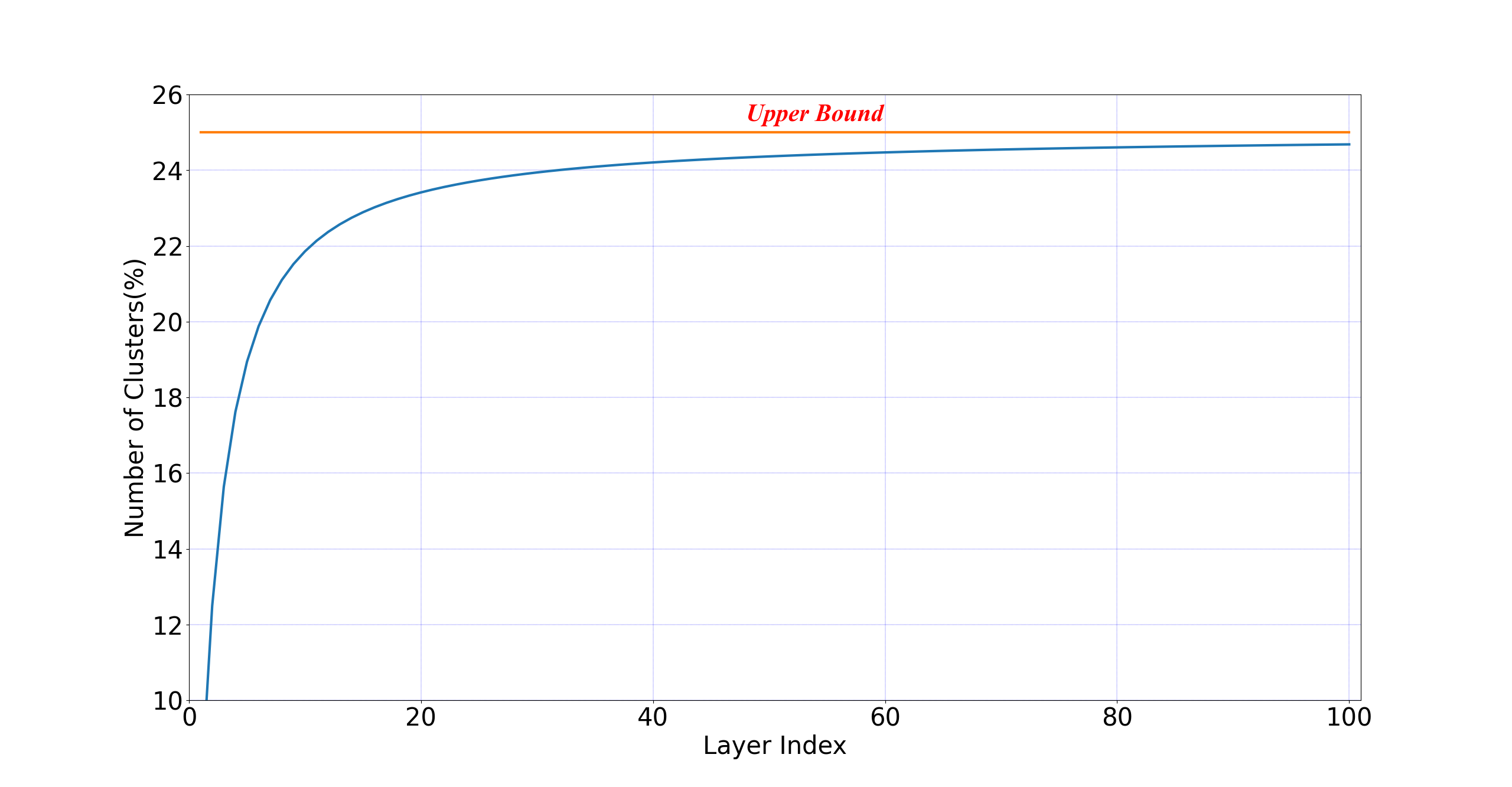}}
\subfigure[Different pruning strategy]{\includegraphics[width=0.495\textwidth]{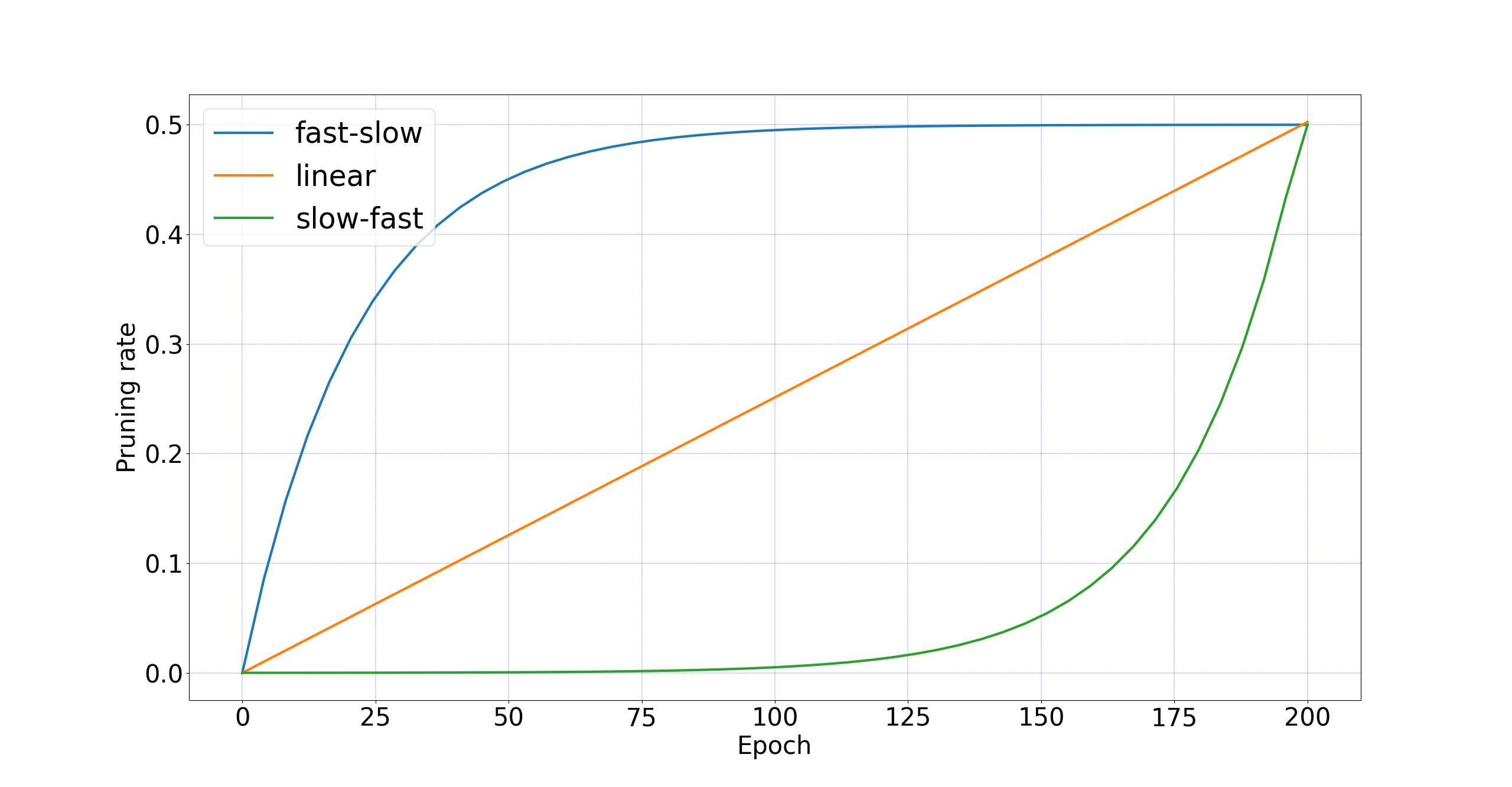}} 
\caption{(a) The adaptive coefficient for clustering, where number of clusters (vertical axis) represents the percentage of total number of filters of each layer. (b) Comparison of different strategies to control the speed of pruning ResNet-56 on CIFAR-10 dataset, where the goal pruning rate is 0.5.} 
% \vspace{-1ex}
\label{fig3}
\end{figure}

\emph{\textbf{Guideline 2: Constant over layers.}} Although Silhouette Coefficient can be precise, such guideline can be quite time-consuming when $N_{out}^{i}$ is large. Besides, if the layer extracts a variety of features, the number of clusters can be massive. This will do harm to the total pruning rate since we need to keep one filter of each cluster, which will be illustrated later in our channel addition operation. In practice, we find that setting $c_i$ to a constant across all layers could perform relatively well for shallow and plain networks (e.g., VGGNet). 

\emph{\textbf{Guideline 3: Adaptive coefficient.}} However, for a deep and complex structure (e.g. ResNet-50), the above two guidelines can be less effective. As the forward propagation goes, the deeper layers suppose to have more channels to extract various features, which logically should be assigned more clusters. We scale the pruning rate to the worse case and restrict it to satisfy the realistic pruning rate $P_r$ as in Eq. \ref{eq12}, which factually treats $P_r$ as the lower pruning bound.  In our experiment, since the pruning rate is same across layers, we can simply scale the arctangent function to modify its upper bound, as shown in Fig. \ref{fig3}(a).

\begin{equation}\label{eq12}
\begin{split}
P_r\leqslant \frac{\sum_{i=1}^L{\left( P_i-\max \left( \frac{c_i}{N_{out}^{i}} \right) \right) N_{out}^{i}}}{\sum_{i=1}^L{N_{out}^{i}}}&\leqslant \frac{\sum_{i=1}^L{\left( P_i-\frac{c_i}{N_{out}^{i}} \right) N_{out}^{i}}}{\sum_{i=1}^L{N_{out}^{i}}},
\\
then \quad \max \left( \frac{c_i}{N_{out}^{i}} \right) &\leqslant \frac{\sum_{i=1}^L{\left( P_i-P_r \right) N_{out}^{i}}}{\sum_{i=1}^L{N_{out}^{i}}}
\end{split}
\end{equation}

\textbf{Filter Selection:} We adopt the Euclidean distance to evaluate the similarity of filters as Eq. \ref{eq6}. In general, those filters close to assigned cluster centroids would have more redundancy. Based on this understanding, we can find the “most similar” filters and set their values to centroids which they belong to.

\begin{align}\label{eq6}
{W_{ij^{*}} {\in} \mathop {arg\min} \limits_{j\in \left[ 1,N_{out}^{i} \right]}\left\| \left. W_{ij}-\mu _{c_{i}^{\left( j \right)}} \right\| \right. _2},
\\
{W_{ij^{*}} = \mu _{c_{i}^{\left( j^* \right)}},\ \text{for}\ 1\leqslant j^*\leqslant N_{out}^{i}*P_i}. \notag
\end{align} 

To avoid severe information loss, we limit ${P_i}$ to gradually increase from the initial value towards the goal pruning rate ${P_{goal}}$. The definition of ${P_i}$ is list as follows:

\begin{equation}
{P_i=f\left( P_{i}^{init},P_{i}^{goal},epoch \right)  \label{eq7}},
\end{equation}
where $P_{i}^{init}$ represents the initial pruning rate for the $i$-th layer. To stabilize the pruning process, we consider two kinds of strategies, which are linear increase and exponential increase respectively.

The linear increase strategy can be written as:

\begin{equation}
{P_i=\frac{P_{goal}}{epoch_{\max}} \times epoch  \label{eq8}},
\end{equation}
where ${epoch_{max}}$ is the number of total training epochs. Starting from zero, the pruning rate will linearly increase until it reach the goal pruning rate, as illustrated in Fig. \ref{fig3}(b). Similarly, the exponential increase strategy can be given by:

\begin{equation}
{P_i=k_1 \times \left( e^{k_2 \times epoch}-1 \right) \label{eq9}},
\end{equation}

where $k_{1}$ and $k_{2}$  are two hyper-parameters to control the growth speed of the pruning rate. For different setting of hyper-parameters, the trend of pruning rate can be fast to slow, or vice versa, as shown in Fig. \ref{fig3}. 

\textbf{Filter Recovery:} Filters can be wrongly reconstructed since the cluster centroids are changing along the whole pruning process. Thus after the selection step, we retrain the network for some epochs so that it can recover from the wrong assignment. Furthermore, our method allows the model to keep the same capacity, which ensures the high performance of the model. As the pruning step is integrated into the normal training schema, the model can be trained and pruned synchronously.

\textbf{Channel Addition:} Before convergence, ASCP iterates over the clustering, selection and recovery procedures. When the model converges, it contains lots of identical filters as redundancy, where we can apply channel addition operation to obtain a compact model. As shown in Fig. \ref{fig2}, suppose filters $W_{i,m}$ and $W_{i,n}$ are identical, we can safely prune one of them by adding corresponding channels of the filters in the next layer, which can be represented by:

\begin{equation} \label{eq10}
\begin{split}
&W_{i+1}*I_{i+1}=W_{i+1}^{'}*I_{i+1}^{'},
\\
&\text{s.t.}\quad  \left\{\begin{array}{lc}
W_{i,m}=W_{i,n}, \\
W_{i+1,:,m}^{'}=W_{i+1,:,m}+W_{i+1,:,n}, \\
I_{i+1}^{'}=W_{i,n}^{C}*I_i, \\
\end{array}\right.
\end{split}
\end{equation}
where $I_{i+1}^{'}$ denotes the input feature map of $i$+1-th layer after pruning $W_{i,n}$. $W_{i+1,:,m}$ and $W_{i+1,:,n}$ represent the $m$-th and $n$-th channel of $W_{i+1}$, respectively. $W_{i,n}^{C}$ is the complementary set of $W_{i,n}$. Although the parameters of batch normalization are ignored, we can simply eliminate this effect by fine-tuning for a few epochs. Therefore, there will be no influence to remove these filters as well as the corresponding feature maps. The details of our ASCP are illustrated in Algorithm \ref{alg1}.

\begin{algorithm}\label{alg1}
% 	\textsl{}\setstretch{1.8}
	\renewcommand{\algorithmicrequire}{\textbf{Input:}}
	\renewcommand{\algorithmicensure}{\textbf{Output:}}
	\caption{ASCP Algorithm}
	\begin{algorithmic}[1]
		\REQUIRE training set \textbf{$D$}, pruning rate $P_{i}$, number of clusters $c_{i}$, model with parameters $W=\left\{ W_i,0\leqslant i\leqslant L \right\}$, total epochs $t_{epoch}$

		\FOR {$epoch=1;epoch\leqslant t_{epoch};epoch++$}
		\STATE Update the model parameter $W$ based on $D$;
		\FOR {$i=1;i\leqslant L,i++$}
		\STATE Filter Clustering based on $c_{i}$;
		\STATE Filter Selection base on Equation \ref{eq6};
		\STATE Set each selected filter $W_{ij^{*}}$ to centroid $\mu _{c_{i}^{\left( j^* \right)}}$ it belongs to;
		\ENDFOR
		\ENDFOR
		\STATE Obtain the pruned model by Channel Addition;
		\ENSURE  The compact model with parameter $W^{'}=W^{t_{epoch}}$
	\end{algorithmic}  
\end{algorithm}

\begin{table}[htbp]
\centering
\caption{Pruning results of ResNet on CIFAR-10. \protect\footnotemark[1]} \label{table1}
\begin{tabular}{c|ccccc} 
\hline
Depth                & Method        & Baseline(\%)   & Pruned Accu.(\%) & Accu. Drop(\%) & Pruned FLOPs(\%)  \\ 
\hline
\multirow{5}{*}{20}  & SFP~\cite{ref10}          & 92.20          & 91.20            & 1.00           & 29.3              \\
                     & ASFP~\cite{ref24}         & 92.89          & 91.62            & 1.27           & 29.3              \\
                     & SRFP~\cite{ref11}         & 92.32          & 91.47            & 0.85           & 29.3              \\
                     & GHFP~\cite{ref26}         & 92.89          & 92.14            & 0.75           & 29.3              \\
                     & \textbf{Ours} & \textbf{92.17} & \textbf{91.73}   & \textbf{0.44}  & \textbf{40.8}     \\ 
\hline\hline
\multirow{6}{*}{56}  & SFP~\cite{ref10}          & 93.59          & 92.26            & 1.33           & 52.6              \\
                     & ASFP~\cite{ref24}         & 93.59          & 93.12            & 0.47           & 52.6              \\
                     & SRFP~\cite{ref11}         & 93.66          & 92.67            & 0.99           & 52.6              \\
                     & FPGM~\cite{ref13}         & 93.59          & 92.93            & 0.66           & 52.6              \\
                     & GHFP~\cite{ref26}         & 94.85          & 93.84            & 1.01           & 52.6              \\
                     & \textbf{Ours} & \textbf{93.59} & \textbf{92.98}   & \textbf{0.61}  & \textbf{60.8}     \\ 
\hline\hline
\multirow{5}{*}{110} & SFP~\cite{ref10}          & 93.68          & 92.90            & 0.78           & 52.3              \\
                     & ASFP~\cite{ref24}         & 94.33          & 93.52            & 0.81           & 52.3              \\
                     & ASFRP~\cite{ref11}        & 94.33          & 93.66            & 0.67           & 52.3              \\
                     & GHFP~\cite{ref26}         & 94.76          & 94.38            & 0.38           & 52.3              \\
                     & \textbf{Ours} & \textbf{94.17} & \textbf{93.81}   & \textbf{0.36}  & \textbf{61.5}     \\
\hline
\end{tabular}
\end{table}

\section{Experiment}

\subsection{Experimental Settings}

% \subsubsection{Datasets}
\textbf{Datasets.} We conduct experiments on two publicly available datasets CIFAR-10 \cite{ref28} and ImageNet \cite{ref29} to show the efficiency of our method. CIFAR-10 dataset contains 60000 32*32 images with 10 classes, 50000 images for training while the remaining for testing. ImageNet is a large-scale dataset which contains 1.28 million 224*224 training images and 50k validation images drawn from 1000 categories. For both datasets, we first preprocess the data by subtracting the mean and dividing the standard-deviation, then adopt the same data augmentation scheme as \cite{ref30} \cite{ref31}. 

% \subsubsection{Network Architecture}
\textbf{Network Architecture.} We study the performance on various mainstream CNN models, including VGGNet \cite{ref32} with a plain structure and ResNet series \cite{ref33} with residual blocks. For CIFAR-10 dataset, we test our ASCP on VGGNet, ResNet-20, 56 and 110, while on ImageNet, we test it on ResNet-18, 34 and 50. Since ResNet is less redundant than VGGNet, we will focus more on it.

% \subsubsection{Configurations}
\textbf{Configurations.} The models are trained from scratch using SGD with a weight decay of 0.0005 and Nesterov momentum of 0.9 \cite{ref34}. For CIFAR-10, we train the model with a batchsize of 128 for 200 epochs and use an initial learning rate as 0.1, while on ImageNet, the number of epochs, batchsize and the initial learning rate are set to 100, 256 and 0.1, respectively. The learning rate is divided by 5 at epoch 60, 120, 160 on CIFAR-10, and is divided by 10 every 30 epochs on ImageNet. We prune all the convolutional layers of VGGNet for it has a plain structure. While for ResNet, due to the existence of shortcut, we prune all the convolutional layers except for the last one of every residual block for simplification. The pruning operation is conducted after every training epoch with the same pruning rate across all selected layers. We compare our results with other state-of-the-art filter pruning methods, e.g., SFP\cite{ref10}, ASFP\cite{ref24}, SRFP/ASFRP\cite{ref11}, FPGM\cite{ref13}, GHFP\cite{ref26}, AutoPrune\cite{ref35}, VCNNP\cite{ref36}, GAL\cite{ref37}, GS\cite{ref38}, Hrank\cite{ref39}, and all the results are obtained from the original papers. We conduct each experiment for 3 times and report the mean value for comparison.

\footnotetext[1]{For the sake of fairness, the baselines are compared with the favorable results in the original papers.}

\begin{table}
\caption{Pruning results of VGG-16 on CIFAR-10. Numbers in parentheses denote the pruning rate. }\label{table2}
\centering
\resizebox{\linewidth}{!}{
\begin{tabular}{c|c|c|c|c} 
\hline
Method             & Baseline(\%)   & Pruned Accu.(\%) & Accu. Drop(\%) & Pruned FLOPs(\%)  \\ 
\hline
AutoPrune~\cite{ref35}          & 92.40          & 91.50            & 0.9            & 23.0              \\
\textbf{Ours(0.4)} & \textbf{93.87} & \textbf{93.92}   & \textbf{-0.05} & \textbf{37.1}     \\
VCNNP~\cite{ref36}              & 93.25          & 93.18            & 0.07           & 39.1              \\
GAL~\cite{ref37}                & 93.96          & 90.73            & 3.23           & 45.2              \\
GS~\cite{ref38}                 & 94.02          & 93.59            & 0.43           & 60.9              \\
\textbf{Ours(0.6)} & \textbf{93.87} & \textbf{93.51}   & \textbf{0.36}  & \textbf{61.9}     \\
Hrank~\cite{ref39}              & 93.96          & 92.34            & 1.62           & 65.3              \\
\textbf{Ours(0.7)} & \textbf{93.87} & \textbf{92.72}   & \textbf{1.15}  & \textbf{72.2}     \\
\hline
\end{tabular}}
\end{table}

\subsection{Pruning Results on CIFAR-10}

\begin{table*}
\centering
\caption{Pruning results of ResNet on ImageNet, where “BL” and "PR" denote baseline and pruned, respectively. }\label{table3}
\begin{tabular}{c|cccccc} 
\hline
Depth               & Method        & \begin{tabular}[c]{@{}c@{}}Top-1 Accu.\\BL/PR(\%)\end{tabular} & \begin{tabular}[c]{@{}c@{}}Top-5 Accu.\\BL/PR(\%)\end{tabular} & \begin{tabular}[c]{@{}c@{}}Top-1 Accu. \\Drop(\%)\end{tabular} & \begin{tabular}[c]{@{}c@{}}Top-5 Accu.\\Drop(\%)\end{tabular} & \begin{tabular}[c]{@{}c@{}}Pruned \\FLOPs(\%)\end{tabular}  \\ 
\hline
\multirow{4}{*}{18} & SFP~\cite{ref10}          & 70.28/67.10                                                    & 89.63/87.78                                                    & 3.18                                                           & 1.85                                                          & 41.8                                                        \\
                    & ASFP~\cite{ref24}         & 70.23/68.02                                                    & 89.51/88.19                                                    & 2.21                                                           & 1.32                                                          & 41.8                                                        \\
                    & ASRFP~\cite{ref11}        & 70.23/67.25                                                    & 89.51/87.59                                                    & 2.98                                                           & 1.92                                                          & 41.8                                                        \\
                    & \textbf{Ours} & \textbf{69.74/67.76}                                           & \textbf{89.18/87.95}                                           & \textbf{1.98}                                                  & \textbf{1.23}                                                 & \textbf{45.6}                                               \\ 
\hline
\multirow{5}{*}{34} & SFP~\cite{ref10}          & 73.92/71.83                                                    & 91.62/90.33                                                    & 2.09                                                           & 1.29                                                          & 41.1                                                        \\
                    & ASFP~\cite{ref24}         & 73.92/72.53                                                    & 91.62/91.04                                                    & 1.39                                                           & 0.58                                                          & 41.1                                                        \\
                    & ASRFP~\cite{ref11}        & 73.92/71.39                                                    & 91.62/90.23                                                    & 2.53                                                           & 1.39                                                          & 41.1                                                        \\
                    & FPGM~\cite{ref13}         & 73.92/71.79                                                    & 91.62/90.70                                                    & 2.13                                                           & 0.92                                                          & 41.1                                                        \\
                    & \textbf{Ours} & \textbf{73.15/71.96}                                           & \textbf{91.15/90.58}                                           & \textbf{1.19}                                                  & \textbf{0.57}                                                 & \textbf{45.1}                                               \\ 
\hline
\multirow{4}{*}{50} & SFP~\cite{ref10}          & 76.15/74.61                                                    & 92.87/92.06                                                    & 1.54                                                           & 0.87                                                          & 41.8                                                        \\
                    & ASFP~\cite{ref24}         & 76.15/75.53                                                    & 92.87/92.73                                                    & 0.62                                                           & 0.14                                                          & 41.8                                                        \\
                    & SRFP~\cite{ref11}         & 76.15/75.98                                                    & 92.87/92.81                                                    & 0.17                                                           & 0.06                                                          & 41.8                                                        \\
                    & \textbf{Ours} & \textbf{75.56/75.13}                                           & \textbf{92.70/92.55}                                           & \textbf{0.43}                                                  & \textbf{0.15}                                                 & \textbf{48.5}                                               \\
\hline
\end{tabular}
\end{table*}

\textbf{ResNet-20/56/110.} We conduct our experiment on ResNet-20, 56, 110 for CIFAR-10 dataset. After pruning the pre-trained model, it is finetuned to recover performance. We summarize the results in Table \ref{table1}, our ASCP achieves comparable performance on CIFAR-10 compared with other filter pruning methods. For example, ASFP accelerates ResNet-56 by 52.6$\%$ with 1.20$\%$ accuracy drop while our ASCP can prune 60.8$\%$ of total FLOPs with only 0.61$\%$ accuracy drop. Notably, when the pruning rate is relatively small, these pruning techniques can not make a big gap with each other. However, as the pruning rate becomes larger, ASCP outperforms with only 0.36$\%$ accuracy loss and 61.5$\%$ of total FLOPs pruned on ResNet-110. This is because ASCP keeps the similar features instead of forcing them towards zeros as other SFP based methods, which demonstrated our ASCP can well protect feature extraction ability of the pruned model.

\textbf{VGG-16.} Table \ref{table2} shows the performance of different pruning techniques on VGG-16. We set the number of clusters to 25$\%$ of the total amount of filters in each layer to prune the pre-trained model, and then varies the pruning rate between 0.4/0.6/0.7 for better comparison. Our ASCP provides lower accuracy loss compared to AutoPrune and VCNNP (-0.05$\%$ v.s. 0.07$\%$, 0.9$\%$) with similar FLOPs reduction. Compared with GAL and Hrank, ASCP outperforms each of them in all aspects (45.2$\%$ v.s. 61.9$\%$ and 65.3$\%$ v.s. 72.2$\%$ in FLOPs reduction, 0.36$\%$ v.s. 3.23$\%$ and 1.15$\%$ v.s. 1.62$\%$ in accuracy loss), which proves its ability to compress and accelerate a neural network with a plain structure.
% \vspace{-1ex}

\subsection{Pruning Results on ImageNet} 
% \vspace{-1ex}
For the ImageNet dataset, we evaluate our ASCP on ResNet-18, 34 and 50. We do not prune the last layer of each residual block and the projection shortcuts for simplification. For ResNet-50, since it has the bottleneck structure which contains three convolutional layers in each block, the actual pruned FLOPs will be larger than ResNet-18 and 34 under same pruning rate. Table \ref{table3} shows the performance of our ASCP compared with previous mentioned methods. As a result, our method can achieve better FLOPs reduction and lower accuracy drop, for instance, our ASCP can prune 48.5$\%$ FLOPs of ResNet-50 with only 0.43$\%$ top-1 and 0.15$\%$ top-5 accuracy drop. 

\subsection{Ablation Study}
\begin{figure}[t]
\centering
\subfigure[Different pruning rate on ResNet-20]{\includegraphics[width=0.495\textwidth]{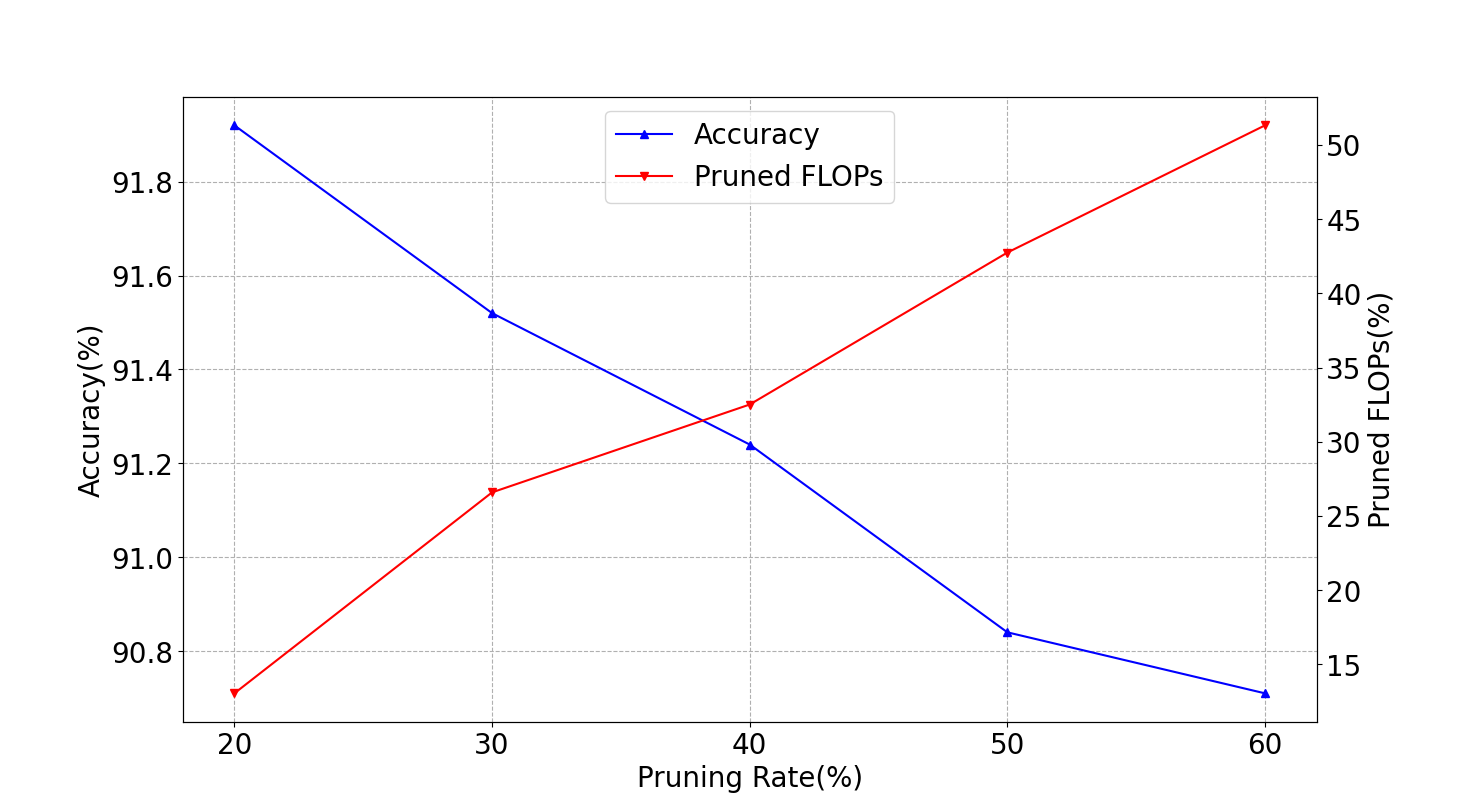}} 
\subfigure[Different number of clusters on ResNet-20]{\includegraphics[width=0.495\textwidth]{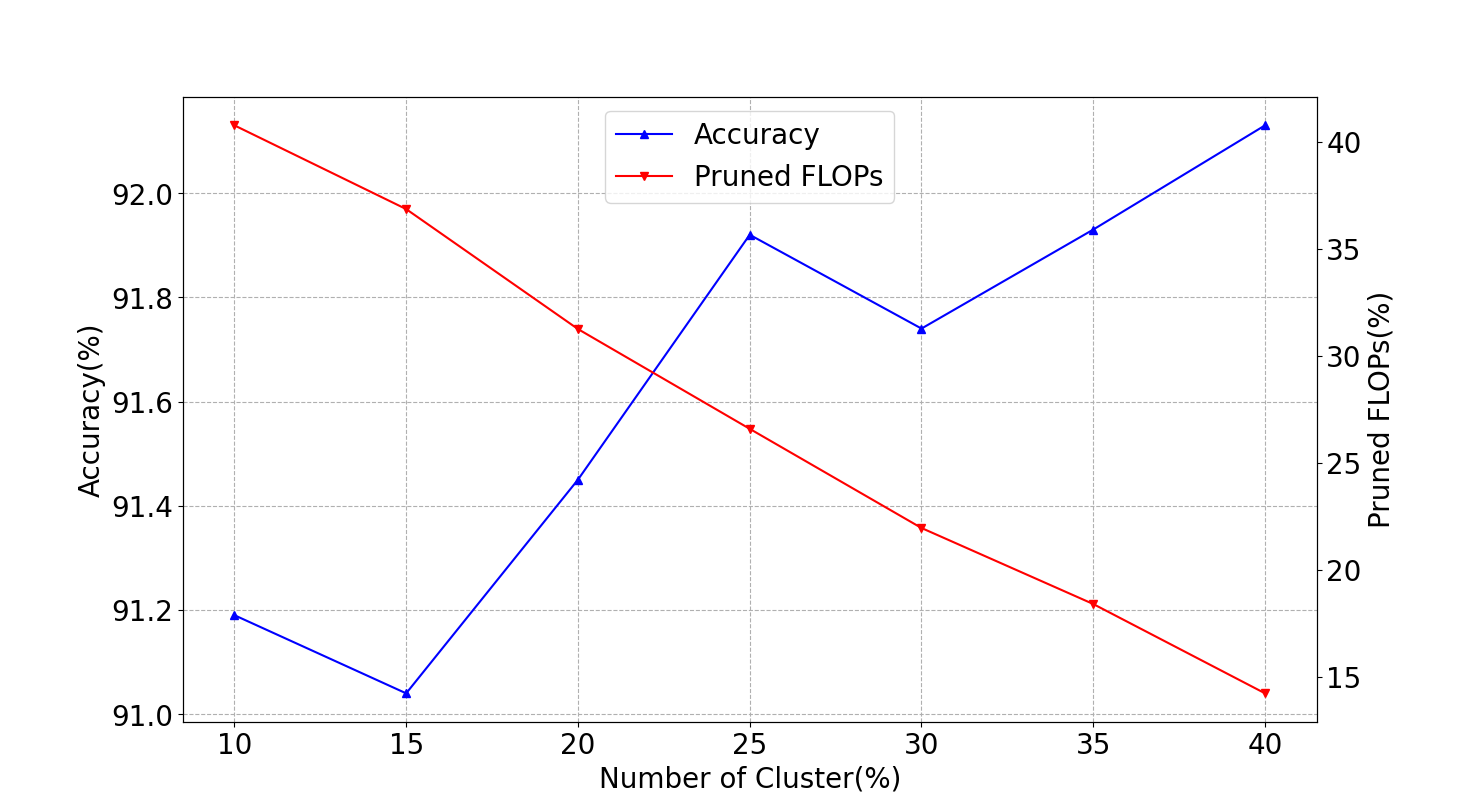}}
\caption{Comparison of test accuracies of different pruning rates and number of clusters for ResNet-20 on CIFAR-10 dataset with the fast-slow stratgy.} 
\vspace{-3ex}
\label{fig4}
\end{figure}

\begin{figure}[t]
\centering
\subfigure[ResNet-20 from pre-train]{\includegraphics[width=0.495\textwidth]{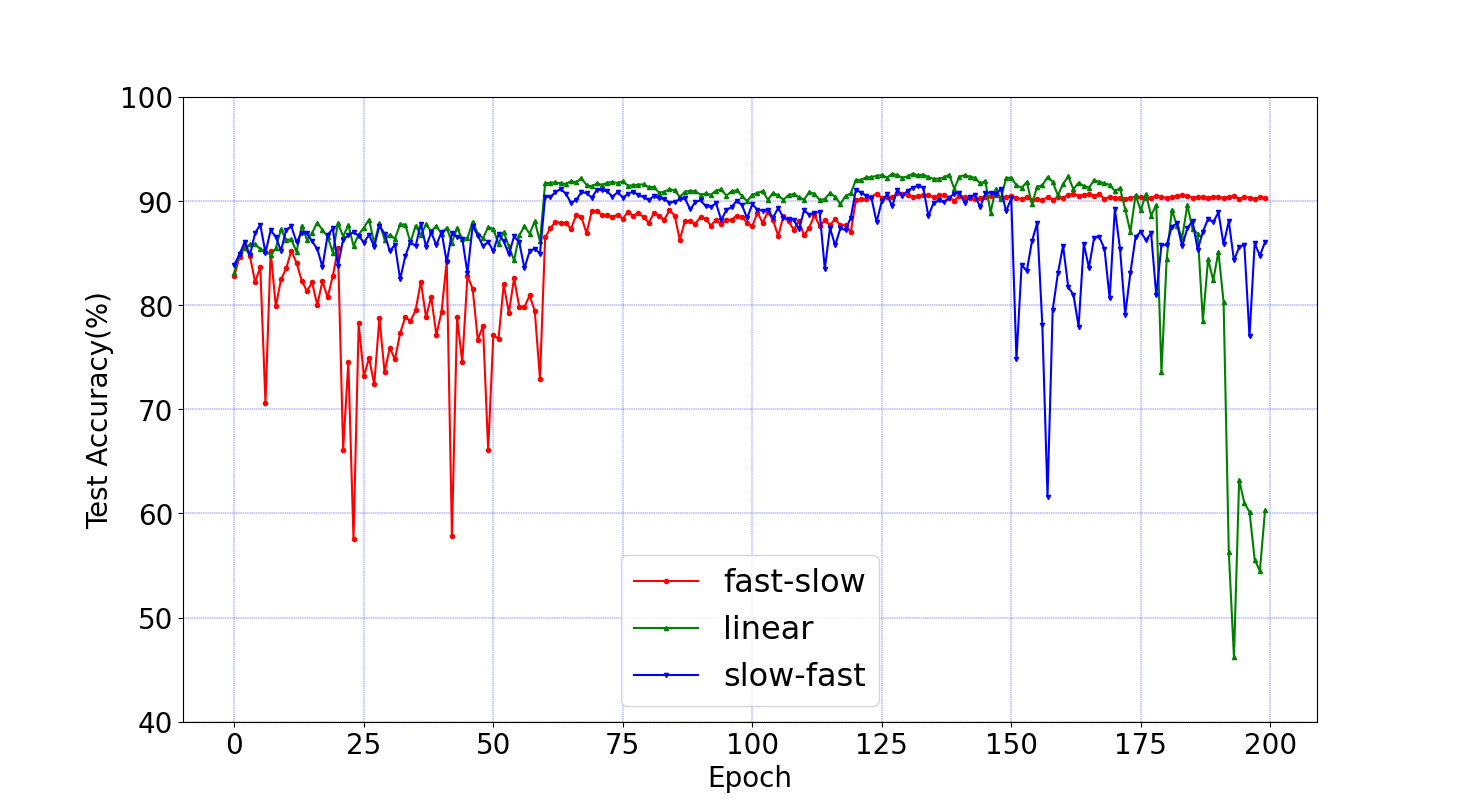}}
\subfigure[ResNet-20 from scratch]{\includegraphics[width=0.495\textwidth]{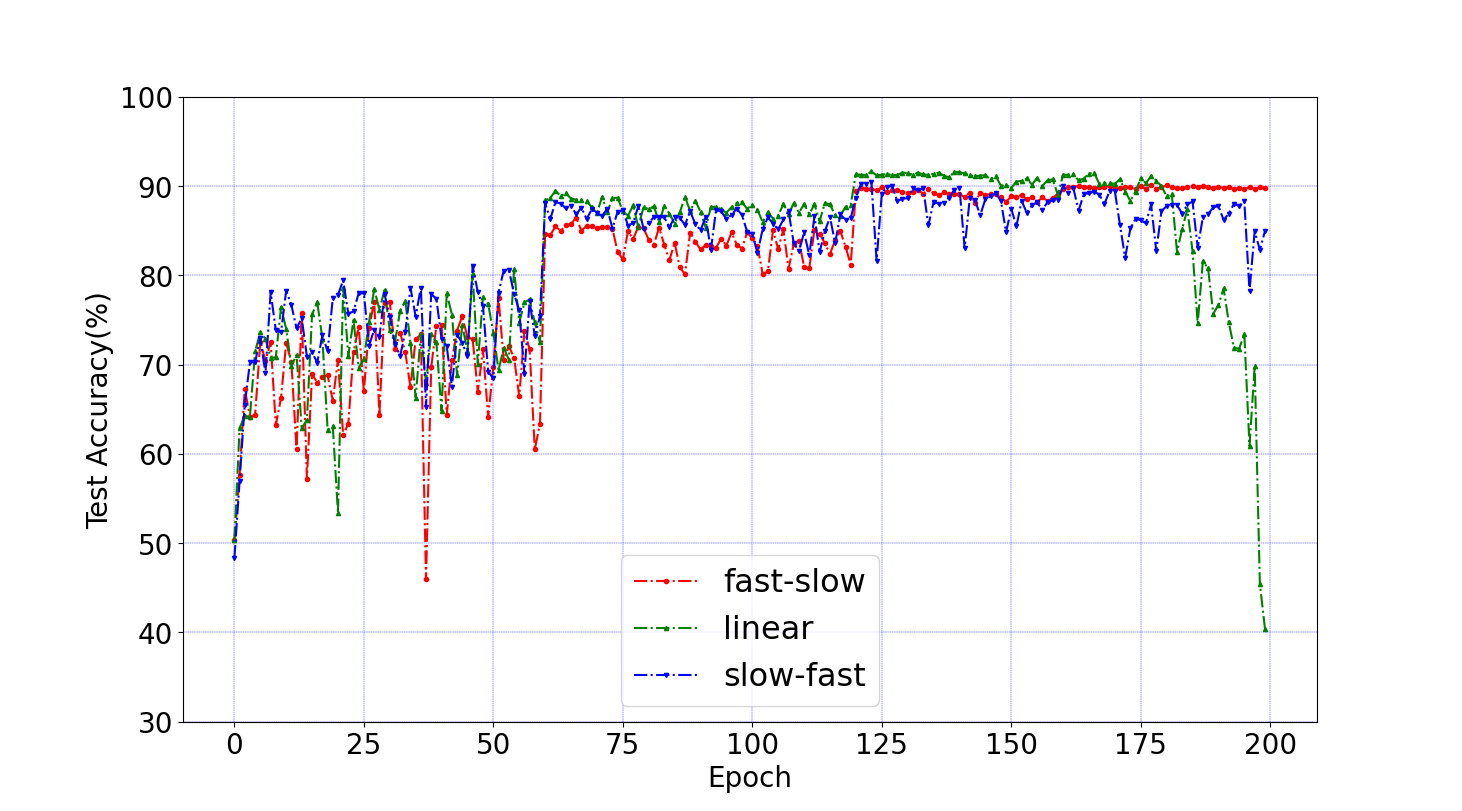}}
\caption{The training process of ResNet-20 on CIFAR-10 with different strategies of pruning rate while the goal pruning rate is set to 50$\%$. Top-1 accuracy is reported for comparison.} 
\vspace{-3ex}
\label{fig5}
\end{figure}
% \vspace{-5ex}

\textbf{Varying pruning rates.} To better shed light on the performance of our ASCP, we present the test accuracy under different pruning rate for ResNet-20 in Fig. \ref{fig4}(a). The network is trained from scratch with the number of clusters $c_{i}$ is fixed at 25$\%$ without fine-tuning. As shown in Fig. \ref{fig4}(a), with the decrease of pruning rate, the test accuracy increases linearly, and the pruned FLOPs decreases gradually as well. When the pruning rate is larger than 50$\%$, the accuracy loss compared with baseline is more than 1$\%$, which is difficult to recover even after fine-tuning. Thus, for a better trade-off between model complexity and performance, the pruning rate should be chosen carefully.

\textbf{Influence of the number of clusters.} In order to explore the influence of $c_{i}$, we compare the test accuracy of pruned ResNet-20 with different $c_{i}$ on CIFAR-10 dataset. As shown in Fig. \ref{fig4}(b), the overall trend of accuracy is that the larger $c_{i}$ is, the higher test accuracy becomes. However, the pruned FLOPs is also downward and there exists some knee points of accuracy, which suggests that $c_{i}$ should not be too large and it needs to be well designed to balance the general performance.

\textbf{Different pruning strategies.} We compare the results of three decay strategies when pruning ResNet-20 on CIFAR-10 with the training epochs increasing. The hyper-parameters $P_{i}$ and $c_{i}$ are 0.8 and 25$\%$, respectively. As shown in Fig. \ref{fig5}, no matter if it is trained from scratch or not, the \textbf{fast-slow} strategy outperforms others although it causes relatively larger accuracy loss at the initial stage of training process. The \textbf{linear} strategy creates identical filters in a much smoother manner which leads to a disastrous non-convergence issue, so it's more efficient to use the \textbf{fast-slow} strategy while pruning.
\vspace{-1ex}
\section{Conclusion}
\vspace{-1ex}
To conclude, in this paper, we point out limitations of previous works and propose a novel filter pruning method based on clustering, named ASCP, to accelerate the deep CNNs. Specifically, ASCP considers the similarity between filters as redundancy and removes them in a softer way without information loss. Thanks to these, ASCP allows filters to be pruned more stable as the training procedures run and achieves the state-of-the-art performance in several benchmarks. In the future, we plan to work on how to combine ASCP with other norm-based criteria and more importantly, other acceleration algorithms, e.g., knowledge distillation and matrix decomposition, to push the performance to a higher stage.

\section{Acknowledgements}
This work was supported in part by the National Key R\&D Program of China (No. 2018YFB1201500), the National Natural Science Foundation of China under Grant No. 61771072.

\bibliographystyle{abbrv}{\small
}

\end{document}